\documentclass[letterpaper, 10 pt, conference]{ieeeconf}  

\IEEEoverridecommandlockouts                              

\usepackage{bm}
\usepackage{graphics} 
\usepackage{epsfig} 
\usepackage{times} 
\usepackage{amsmath} 
\usepackage{amssymb}  
\usepackage{cite}
\usepackage{hyperref}
\usepackage{subcaption}
\hypersetup{
    colorlinks=true,
    linkcolor=blue,
    urlcolor=blue,
    citecolor=blue
}

\usepackage{algorithm}
\usepackage{algorithmic}

\usepackage{nomencl}
\makenomenclature
\renewcommand{\nomname}

\begin{document}

\pagestyle{empty}

\clearpage
\begin{center}
    \large \textit{Published in IEEE/RSJ International Conference on Intelligent Robots and Systems (IROS) 2025} 
    \vspace{3em}
\end{center}

\parbox{0.9\textwidth}{ 

    \noindent \textbf{Copyright Statement}
    
    \vspace{1em}
    © 2025 IEEE. Personal use of this material is permitted. Permission from IEEE must be obtained for all other uses, including reprinting/republishing this material for advertising or promotional purposes, creating new collective works, for resale or redistribution to servers or lists, or reuse of any copyrighted component of this work in other works.
}
\vspace{3em}

\parbox{0.9\textwidth}{
\textbf{Citation and Access Information}

\vspace{1em}
F. Jing et al., "Steady-State Drifting Equilibrium Analysis of Single-Track Two-Wheeled Robots for Controller Design," 2025 IEEE/RSJ International Conference on Intelligent Robots and Systems (IROS), Hangzhou, China, 2025, pp. 16515-16522.

\vspace{1em}
DOI: \href{https://doi.org/10.1109/IROS60139.2025.11247283}{10.1109/IROS60139.2025.11247283}

\vspace{1em}
IEEE Xplore: \href{https://ieeexplore.ieee.org/document/11247283}{https://ieeexplore.ieee.org/document/11247283}
}
\clearpage

\title{\LARGE \bf
Steady-State Drifting Equilibrium Analysis of Single-Track Two-Wheeled Robots for Controller Design
}

\author{Feilong Jing$^{\dag,1}$, Yang Deng$^{\dag,1}$, Boyi Wang$^1$, Xudong Zheng$^2$, Yifan Sun$^1$, Zhang Chen$^{\ast,1}$ and Bin Liang$^1$
\thanks{This work was supported in part by National Natural Science Foundation of China under Grants 62203252 and 62073183.} 
\thanks{$^{\dag}$ Feilong Jing and Yang Deng contributed equally to this work.} 
\thanks{$^{\ast}$ Zhang Chen is the correspongding author. } 
\thanks{$^1$ The Department of Automation, Tsinghua University, Beijing, 100084, China. 
{\tt\small (email: \{jfl21, by-wang19, sunyf23\}@mails.tsinghua.edu.cn; \{dengyang, cz\_da, bliang\}@tsinghua.edu.cn).}
}
\thanks{$^2$ Qiyuan Lab, Beijing, 100095, China. {\tt\small (email: zhengxudong@qiyuanlab.com)}}
}
\maketitle
\begin{abstract}

Drifting is an advanced driving technique where the wheeled robot's tire-ground interaction breaks the common non-holonomic pure rolling constraint. This allows high-maneuverability tasks like quick cornering, and steady-state drifting control enhances motion stability under lateral slip conditions. While drifting has been successfully achieved in four-wheeled robot systems, its application to single-track two-wheeled (STTW) robots, such as unmanned motorcycles or bicycles, has not been thoroughly studied. To bridge this gap, this paper extends the drifting equilibrium theory to STTW robots and reveals the mechanism behind the steady-state drifting maneuver. Notably, the counter-steering drifting technique used by skilled motorcyclists is explained through this theory.
In addition, an analytical algorithm based on intrinsic geometry and kinematics relationships is proposed, reducing the computation time by four orders of magnitude while maintaining less than 6\% error compared to numerical methods. Based on equilibrium analysis, a model predictive controller (MPC) is designed to achieve steady-state drifting and equilibrium points transition, with its effectiveness and robustness validated through simulations.

\end{abstract}

\section{Introduction}

Drifting, an advanced driving technique that breaks lateral constraints of a wheeled robot's tires, has been widely adopted by skilled human racers to achieve quick cornering with minimal speed loss, especially in off-road competitions. Although four-wheeled robots have already achieved autonomous drifting \cite{cutler2016autonomous,bhattacharjee2018autonomous,acosta2018teaching,cai2020high,yang2022hierarchical,lu2023consecutive}, to the best of the authors' knowledge, single-track two-wheeled (STTW) robots, usually referred to unmanned motorcycles or bicycles equipped with steering and rear wheel actuators, have not yet accomplished this capability. In comparison, the challenge of STTW robots drifting primarily lies in the additional degree of freedom (DOF) introduced by lateral tilt, which means that the robot must maintain self-balance while sustaining the slip angle. In addition, drifting control is closely related to the safety of STTW robots, as tire slippage is a common phenomenon during rapid acceleration or deceleration, as well as on slippery road surfaces in rain or snow. Therefore, drifting control techniques for STTW robots are crucial for enhancing  both maneuverability and safety. 

\begin{figure}[t]
  \centering
    \begin{subfigure}[t]{1 \linewidth}
    \includegraphics[width = 1.0 \linewidth]{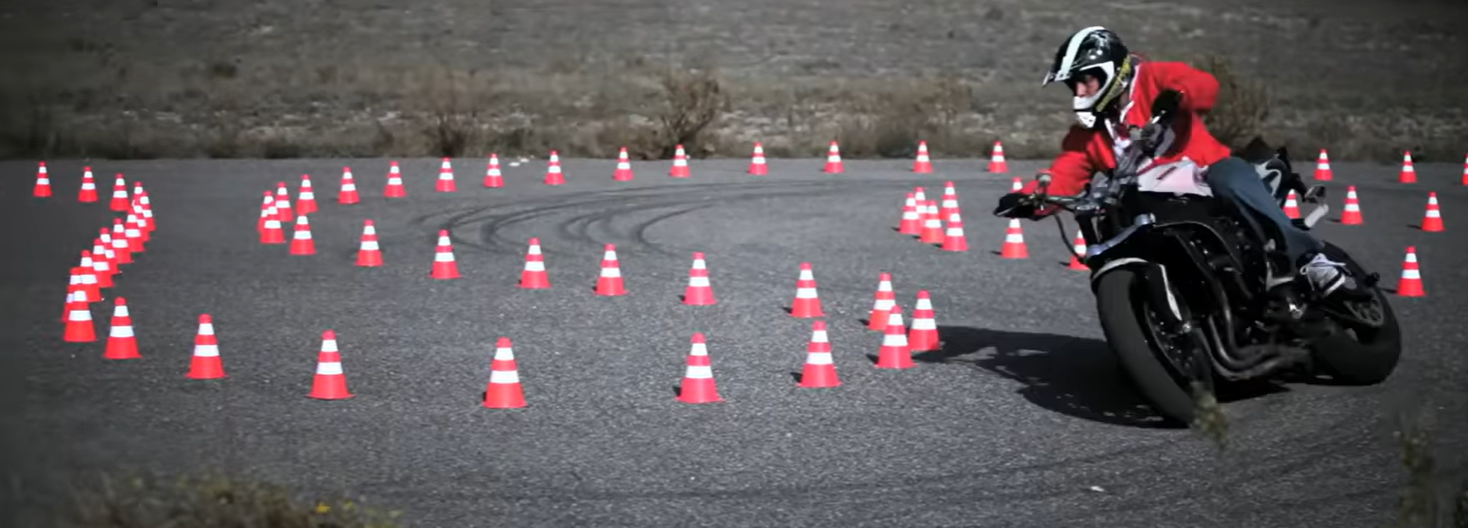}
    \end{subfigure} 
    
    \begin{subfigure}[t]{1 \linewidth}
    \includegraphics[width = 1.0 \linewidth]{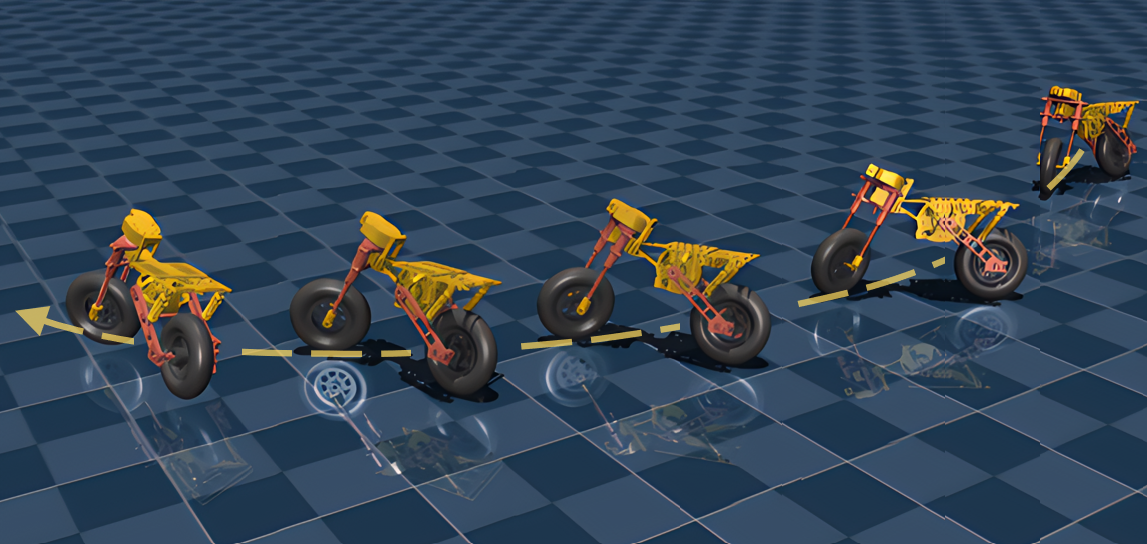}
    \end{subfigure}
    
    \caption{Drifting maneuver of skilled motorcyclists and STTW robots. The top picture shows that motorcyclists use the counter-steering drifting technique \cite{motorcyclist}, and the bottom is snapshots of the STTW robot drifting in MuJoCo simulation.}
    \label{fig: cyclist and mujoco simulation}
\end{figure}

The existing literature on autonomous drifting control has mainly focused on four-wheeled robots, which provides valuable insights for STTW robots. The three-state (yaw rate, lateral velocity, longitudinal velocity) model with nonlinear tire-road friction was used to analyze the unstable drifting dynamics of four-wheeled robots and reveal the existence of the steady-state drift equilibrium \cite{hindiyeh2009equilibrium}. Reference \cite{milani2022vehicle} further used the portrait approach to identify the stability of these equilibria. Based on the drifting equilibrium theory, linear quadratic regulator (LQR) \cite{velenis2011steady,baur2019experimentally} was designed for the drifting control of four-wheeled robots. In addition to steady-state drifting control, tracking controllers were developed to achieve autonomous drifting along a constant-curvature circle \cite{goh2016simultaneous} or a general path \cite{goh2020toward}. These model-based methods exploited drifting equilibrium points to transform the problem into an error stabilization system, which facilitates the design of drifting controllers. In addition, data-driven approaches were also applied to drifting control \cite{cutler2016autonomous,bhattacharjee2018autonomous,acosta2018teaching,cai2020high}, but they provided limited insight into the underlying drifting mechanisms. Notably, drifting equilibrium points were used by Cutler et al. \cite{cutler2016autonomous} to construct the cost function of the learning algorithm, thereby facilitating steady-state drifting.

Due to the inherent instability of STTW robots, equilibrium theory has always been closely tied to the motion control of such type of robot \cite{tian2022steady}. In recent decades, significant developments have been achieved in the balance control issues of STTW robots, including stationary balance control \cite{keo2009controlling,wang2023observer}, constant-speed balance control \cite{defoort2009sliding,yu2018steering}, variable-speed balance control \cite{sun2022fuzzy,wang2023equilibrium1,wang2023equilibrium2}, and trajectory tracking control \cite{yi2006trajectory,he2022learning,chen2022gaussian}. However, the self-balance control under tire slippage is rarely considered in the aforementioned studies, or it is simply treated as a disturbance to the system. According to \cite{cossalter2014motorcycle}, the condition for sideslip balance of STTW robots is that the resultant of the front and the rear wheel forces produces the centripetal component directed towards the turn center. This only provides the force balance condition without further specifying the drifting equilibrium point, making it difficult to design a drifting controller. Yi et al. \cite{yi2009autonomous} proposed an autonomous motorcycle dynamics model with the existence of lateral sliding velocity at each wheel contact point. However, the non-holonomic constraint between the rear wheel velocity and the robot's yaw rate under non-slip conditions is still incorporated into the dynamic equations, making the model unsuitable for large sideslip angles. Zhang et al. \cite{zhang2022non} developed a multibody bicycle model with 24 generalized coordinates, and conducted simulation validation of rear-wheel slipping movement, but the model's reliance on numerical algorithms for solving constraint forces makes it challenging for use in drifting controller. In conclusion, there is a lack of research on deeply revealing the mechanism of the steady-state drifting maneuver of STTW robots, identifying the drifting equilibrium points and realizing stable drifting control.
 
In this paper, we extend the equilibrium theory and controller design to the autonomous drifting of STTW robots. 
Firstly, a dynamic model with large sideslip angle is established by analyzing the forces acting on STTW robot’s center of mass (COM) during drifting. Then, by leveraging the force balance conditions and the relative motion of the rear wheel in steady-state drifting, an analytical solution for the approximated drifting equilibrium point is proposed, significantly speeding up the solution process with a minor trade-off in precision. The drifting mechanisms of STTW robots are explained through the analysis of equilibrium point curves. Moreover, the proposed model also illustrates the counter-steering drifting technique of skilled motorcyclists. Finally, based on the drifting model and equilibria, a model predictive controller (MPC) is applied to achieve steady-state drifting control in MuJoCo \cite{todorov2012mujoco} simulations, which validates the effectiveness of the proposed drifting model. 

The contributions of this work can be summarized as follows:
\begin{itemize}
    \item A dynamical model of STTW robots with large sideslip angles is put forward, revealing the intrinsic relation among the robots' states during steady-state drifting maneuvers.
    \item An analytical drifting equilibrium solving algorithm (ADESA) is proposed to provide solutions without using numerical methods, which is more suitable for embedded implementation on real robots.
    \item This work achieves the autonomous drifting control of STTW robots in simulation, a challenging action which has not been thoroughly studied in the literature.
\end{itemize}

The remainder of this paper is organized as follows. Section II constructs the drifting model of STTW robots. Section III proposes the ADESA algorithm and analyzes the drifting equilibrium curves, and Section IV provides the MPC algorithm for the drifting control of STTW robots. The MuJoCo simulation results are presented and analyzed in Section V. Section VI concludes this paper and provides remarks on future research directions.

\section{Drifting Model}

In this section, we derive the drifting dynamics model of STTW robots using the Newton-Euler method with several assumptions. This model forms the basis for the subsequent drifting equilibrium analysis and serves as the nominal model for the MPC algorithm.

\subsection{Nomenclature}
\nomenclature[01]{\(\quad O$-$XYZ\)}{The world coordinate system.}
\nomenclature[02]{\(\quad o$-$xyz\)}{The body coordinate system with $x$ axis attached to the wheel-base line and $z$ axis pointing downward.}
\nomenclature[03]{\(\quad N_f, N_r\)}{Normal wheel contact forces of the front and rear wheel, respectively.}
\nomenclature[04]{\(\quad f_{ix}, f_{iy}\)}{Friction forces of the front and rear wheel along the $x,y$ axis, $i \in \{f,r\}$.}
\nomenclature[05]{\(\quad a_{rx},a_{ry}\)}{Accelerations of the rear wheel contact point along the $x,y$ axis.}
\nomenclature[06]{\(\quad v_{ix},v_{iy}\)}{Velocities of the front and rear wheel contact point  along the $x,y$-axis, $i \in \{f,r\}$.}
\nomenclature[07]{\(\quad \omega_f,\omega_r\)}{Angular velocities of the front and rear wheel.}
\nomenclature[08]{\(\quad \bm{v}_g, \bm{a}_g\)}{Velocity and acceleration vectors of the robot's COM point $G$ expressed in the world frame.}
\nomenclature[09]{\(\quad \varphi,\psi\)}{The roll and yaw angles of the robot.}
\nomenclature[10]{\(\quad \delta, \delta_f\)}{The steering angle and the projection of the steering angle on the ground.}
\nomenclature[11]{\(\quad \delta_r\)}{Sideslip angle of the real wheel.}
\nomenclature[11]{\(\quad m\)}{Total mass of the robot.}
\nomenclature[12]{\(\quad Ic_{xx},Ic_{zz}\)}{Moment of inertia of the robot with $x$ axis and $z$ axis.}
\nomenclature[13]{\(\quad I_f, I_r\)}{Moment of inertia of the front and rear wheel, respectively.}
\nomenclature[14]{\(\quad a\)}{Horizontal distance between the center of mass of the robot and the rear wheel contact point.}
\nomenclature[15]{\(\quad b\)}{Horizontal distance between the contact points of the front and rear wheels.}
\nomenclature[16]{\(\quad c\)}{Trail length.}
\nomenclature[17]{\(\quad h\)}{Height of the robot's COM.}
\nomenclature[18]{\(\quad r\)}{Wheel radius of the front and rear wheels.}
\nomenclature[19]{\(\quad \lambda\)}{Caster angle.}
\nomenclature[20]{\(\quad \mu\)}{Coulomb friction coefficient of the rear wheel.}

\printnomenclature[1.8cm]

\begin{figure}[H]
  \centering{\includegraphics[width=1\linewidth]{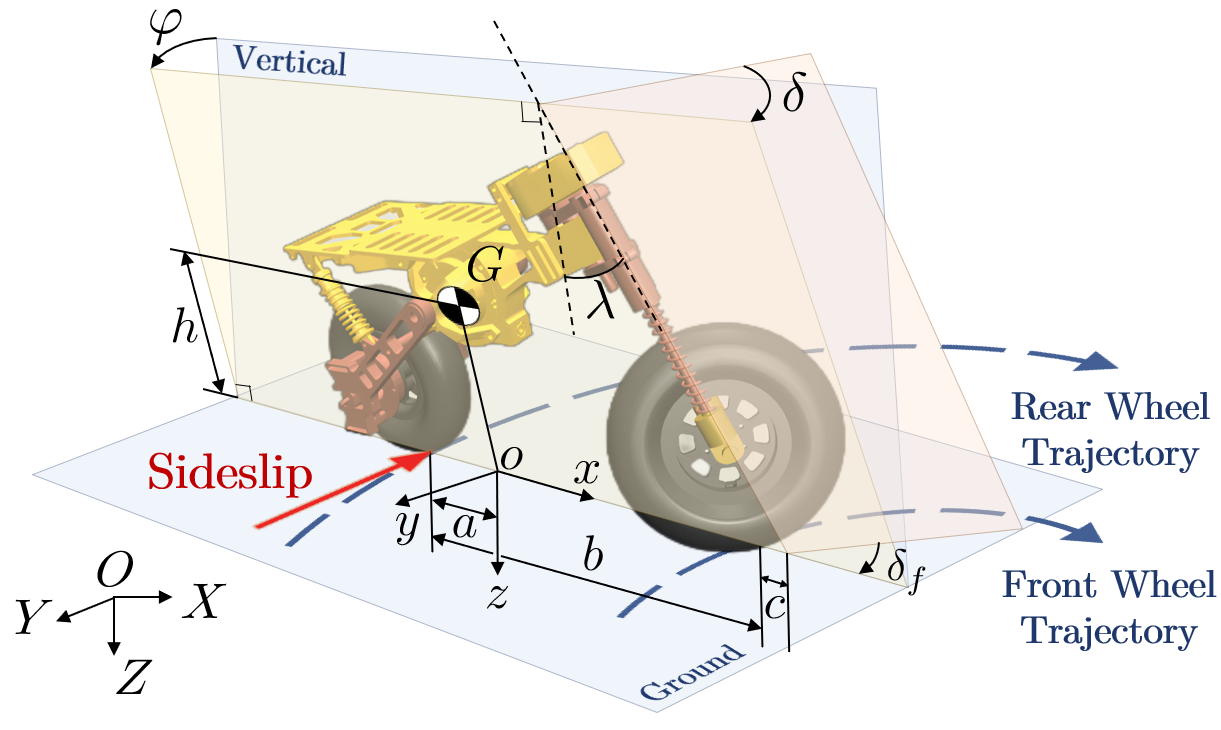}}
  \caption{Schematic of the drifting STTW robot}
  \label{fig: 3d drifting model}
\end{figure}

\subsection{Geometry and kinematics relationships}
Fig. \ref{fig: 3d drifting model} shows a schematic of the drifting STTW robot. The positive directions of the angles, angular velocities, and angular accelerations aforementioned follow the right-hand rule. To simplify the drifting model, we make the following assumptions. 

(i) While the robot is drifting, the front wheel remains pure rolling without lateral slip, whereas the rear wheel breaks the lateral constraint and undergoes side slip. This is similar to the drifting scenario of human drivers. Additionally, we assume that the rear wheel friction follows the Coulomb friction model;

(ii) The thickness of the wheels is neglected so the contact between the wheel and the ground is a single point; 

(iii) The change in the system's overall moment of inertia caused by the steering rotation is negligible.

For simplicity, the trigonometric functions sin($\theta$), cos($\theta$), 
and tan($\theta$) are abbreviated as $s_{\theta}$, $c_{\theta}$, and $t_{\theta}$, respectively.

According to \cite{cossalter2014motorcycle}, the relationship between the steering angle $\delta$ and the projection angle $\delta_f$ is
\begin{equation}
    t_{\delta_f} \, c_{\varphi} = t_{\delta} \, c_{\lambda},
    \label{eq: steering projection}
\end{equation}
which makes it easier to describe the steering angle in the horizontal plane. As we have assumed that the front wheel keeps pure rolling while the STTW robot is drifting, such constraints are derived as
\begin{align}
    v_{rx} + \omega_f r c_{\delta_f} &= 0  \label{eq: constraint fx},\\
    v_{ry} + \dot{\psi}b + \omega_f r s_{\delta_f} &= 0. \label{eq: constraint fy}
\end{align}

To analyze the forces acting on the robot in the COM frame, we first obtain its acceleration expressions. The vector $\bm{r}_{rg}$ from the rear wheel's contact point to point $G$ is expressed in the $o\text{-}xyz$ frame as
\begin{equation}
    \bm{r}_{rg}=\begin{bmatrix}
                 a & hs_{\varphi} & -hc_{\varphi}
                \end{bmatrix}^T.
\end{equation}

Since the $o\text{-}xyz$ frame possesses an angular velocity $[0,\,0,\,\dot{\psi}]^T$ with respect to $O\text{-}XYZ$, the velocity and the acceleration of point G denoted in the world frame are derived as
\begin{align}
    \begin{split}
        \bm{v}_g &= \bm{v}_r + \dot{\bm{r}}_{rg} + [0,\,0,\,\dot{\psi}]^T \times \bm{r}_{rg}, \\
    \end{split} \\
    \begin{split}
        \bm{a}_g &= \dot{\bm{v}}_g + [0,\,0,\,\dot{\psi}]^T \times \bm{v}_g. \\
    \end{split}
\end{align}

\begin{figure}[H]
  \centering{\includegraphics[width=0.82\linewidth]{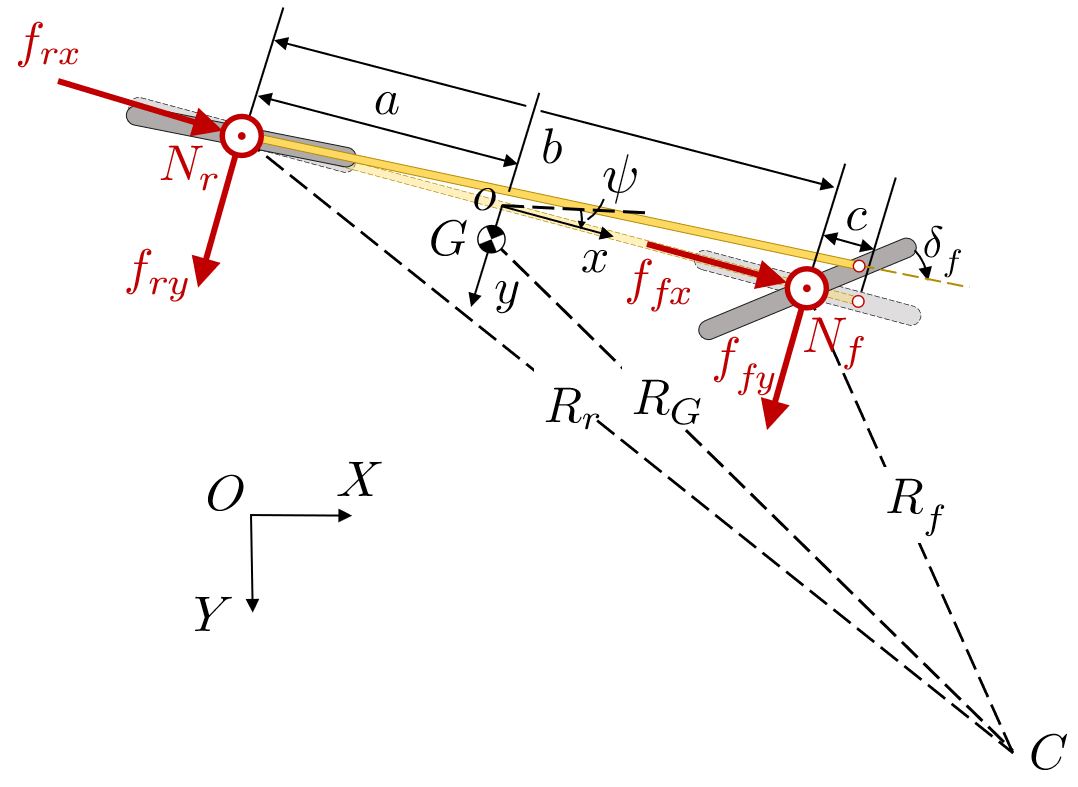}}
  \caption{Top view of the drifting STTW robot}
  \label{fig: top view}
\end{figure}

\subsection{Drifting dynamics}
The dynamics of the drifting STTW robots is derived by the Newton-Euler method which provides a clear representation of the relationship between the forces and the robot's motion. As depicted in Fig. \ref{fig: top view}, in the COM frame, the force balance equations along the $x,y,z$ axis and the moment balance in the pitch direction are expressed as
\begin{align}
    f_{fx} + f_{rx} &= ma_{gx},  \label{eq: x force balance}\\
    f_{fy} + f_{ry} &= ma_{gy},  \label{eq: y force balance}\\
    N_f + N_r + ma_{gz} &= mg,\\
    N_f(b-a) + f_{fx}hc_{\varphi} + f_{rx}hc_{\varphi} &= N_r a,
\end{align}
where $a_{gk}$ ($k\in \{x,y,z\}$) is the component of $\bm{a}_g$ along the $k$-axis. Then, the normal forces are determined by
\begin{align}
    N_f &= \frac{ma(g-a_{gz})-mha_{gx}c_{\varphi}}{b}, \label{eq:Nf}\\
    N_r &= \frac{m(b-a)(g-a_{gz})+mha_{gx}c_{\varphi}}{b}.\label{eq:Nr}
\end{align}

While the STTW robot is drifting, the rear wheel holds a sideslip angle and the rear friction follows the Coulomb friction model, which can be derived as
\begin{align}
    f_{rx}^2 + f_{ry}^2 = \mu N_r^2, \\
    \frac{f_{ry}}{f_{rx}} = \frac{v_{ry}}{v_{rx} + \omega_rr}.
\end{align}

Combined with the force balance equations (\ref{eq: x force balance}) and (\ref{eq: y force balance}), the friction forces of the rear and front wheel are determined by
\begin{align}
    f_{rx} &= -\frac{\mu \left(v_{rx} +\omega_r  r\right) N_r}{\sqrt{\left(v_{rx} +\omega_r  r\right)^2 +v_{ry}^2}}, \label{eq:frx} \\
    f_{ry} &= -\frac{\mu v_{ry} N_r}{\sqrt{\left(v_{rx} +\omega_r  r\right)^2 +v_{ry}^2}}, \\
    f_{fx} &= ma_{gx} - f_{rx}, \\
    f_{fy} &= ma_{gy} - f_{ry}. \label{eq:ffy}
\end{align}

So far, the wheel-ground contact forces have been formulated in terms of the robot's kinematic variables. Then, we can get the equations of roll, pitch, and the front wheel motion:
\begin{align}
    Ic_{xx} \ddot{\varphi} &= (N_f+N_r)hs_{\varphi}  \nonumber \\
    & \qquad\qquad - (f_{ry}+f_{fy})hc_{\varphi} + I_r \omega_r \dot{\psi} c_{\varphi},  \label{eq: eom varphi} \\ 
    Ic_{zz} \ddot{\psi} &= (f_{fx}+f_{rx})hs_{\varphi} + f_{fy}(b-a) - f_{ry}a, \label{eq: eom psi} \\
    I_f \dot{\omega}_f &= f_{fx}rc_{\delta_f} + f_{fy} r s_{\delta_f}. \label{eq: eom wf}
\end{align}

The state and input vectors of the drifting STTW robot model are chosen as
\begin{align}
    \bm{x} &= \begin{bmatrix}
        \delta & \varphi & \dot{\varphi} & \dot{\psi} & \omega_f
    \end{bmatrix}^T, \\
    \bm{u} &= \begin{bmatrix}
        \dot{\delta} & \omega_r
    \end{bmatrix}^T.
\end{align}

By substituting constraint equations (\ref{eq: steering projection})--(\ref{eq: constraint fy}) and the wheel-ground contact force expressions (\ref{eq:Nf})--(\ref{eq:Nr}) and (\ref{eq:frx})--(\ref{eq:ffy}) into the motion equations (\ref{eq: eom varphi})--(\ref{eq: eom wf}), the drifting dynamics model is reformulated in a matrix form as
\begin{equation}
    \begin{bmatrix}
        \bm{I}_{2\times2} & \bm{0}\\
        \bm{0} & \bm{M}_{3\times3}
    \end{bmatrix}\bm{\dot{x}} = 
    \begin{bmatrix}
        \dot{\delta} \\
        \dot{\varphi} \\
        \bm{F}(\bm{x},\bm{u})
    \end{bmatrix}.
    \label{eq: drifting model}
\end{equation}

Typically, the mass matrix $\bm{M}$ depends only on the generalized positions. For instance, in \cite{yi2009autonomous}, $\bm{M}$ is related to the robot's steering and roll angle. However, in the drifting dynamics model (\ref{eq: drifting model}), $\bm{M}$ is related to $(\delta, \varphi, \dot{\psi}, \omega_r,\omega_f)$. This is because in the drifting scenario, the direction of the rear wheel's friction is determined by the wheel-ground relative velocity, which makes $\bm{M}$ dependent on the angular velocity $\omega_r$ and $\omega_f$ as well. The constraint (\ref{eq: constraint fy}) also brings the yaw rate $\dot{\psi}$ into $\bm{M}$. Therefore, the unique drifting constraints lead to a much more complex mass matrix than the standard form.


\section{Drifting Equilibrium}
In this section, we propose two methods for solving drifting equilibrium: the first one is to use numerical techniques to find the roots of (\ref{eq: drifting model}) while the other provides an approximate analytical solution based on geometry and kinematics relationships. By calculating the drifting equilibrium curves, we discuss the underlying drifting mechanisms.

\subsection{Numerical Method}
To obtain the drifting equilibrium, we can simply let $\dot{\bm{x}}=0$ in the drifting dynamics model (\ref{eq: drifting model}) and solve the equations
\begin{equation}
    \begin{bmatrix}
        \dot{\delta} ^{ss}\\
        \dot{\varphi}^{ss} \\
        \bm{F}(\bm{x}^{ss},\bm{u}^{ss})
    \end{bmatrix} = 0,
    \label{eq: eq solve}
\end{equation}
where the superscript $ss$ denotes the steady-state drifting equilibrium values for that variable. Equations (\ref{eq: eq solve}) contain large numbers of trigonometric functions and higher-order terms, so the roots can only be found through numerical methods.

It is obvious that $\dot{\delta}^{ss}=\dot{\varphi}^{ss}=0$ according to (\ref{eq: eq solve}). There are still five variables in $\bm{x}$ and $\bm{u}$ to be solved through $\bm{F}(\bm{x}^{ss},\bm{u}^{ss})=0$, which we define as the equilibrium states, represented by the vector $\bm{\xi}$:
\begin{equation}
    \bm{\xi} = \begin{bmatrix}
        \delta & \varphi & \dot{\psi} & \omega_f & \omega_r
    \end{bmatrix}^T
\end{equation}

Given that $\bm{F}(\bm{x},\bm{u})$ is a $3\times1$ vector, in most cases, a drifting equilibrium point can be determined by specifying the values of any two variables in $\bm{\xi}$.

\subsection{Analytical Method}

\begin{figure}[h]
  \centering{\includegraphics[width=0.9\linewidth]{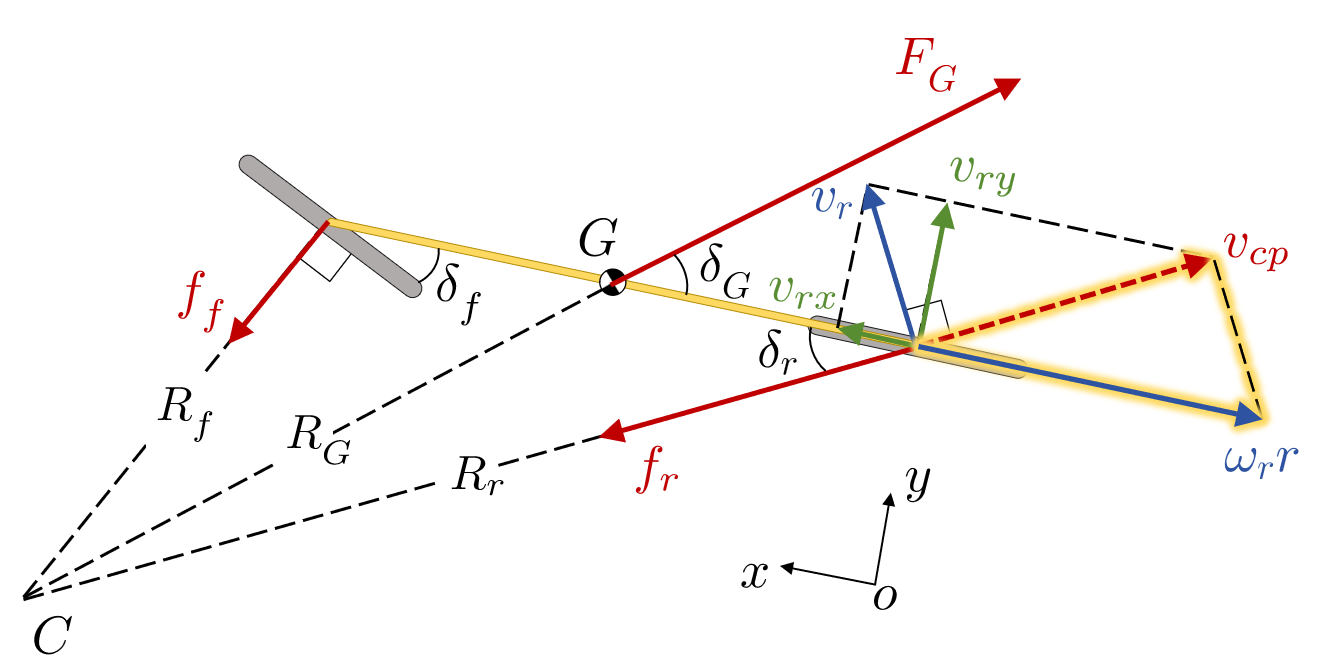}}
  \caption{Simplified drifting model of the STTW robot}
  \label{fig: simplified drifting model}
\end{figure}

Numerical method for solving drifting equilibrium is computationally intensive, making it difficult to implement on robotic embedded systems.
Thus, we propose an approximate analytical solving algorithm to accelerate the equilibrium point computation process based on the geometry and kinematics relationships.

As shown in Fig. \ref{fig: simplified drifting model}, the roll angle $\varphi$ is considered to be a small angle, so the COM point $G$ is approximately located directly above the wheel-base line. At the drifting equilibrium point, the front wheel friction $f_f$ is aligned perpendicular to the direction of front wheel travel. The inertial force $F_G$ acting on the COM is expressed as
\begin{equation}
    F_G = m\dot{\psi}^2R_G,
    \label{eq:Fg}
\end{equation}
where $R_G$ is the distance between point $G$ and the instantaneous center of rotation $C$.

\begin{figure*}[t]
  \centering
    \begin{subfigure}[t]{0.3 \textwidth}
    \includegraphics[width = 1.0 \textwidth]{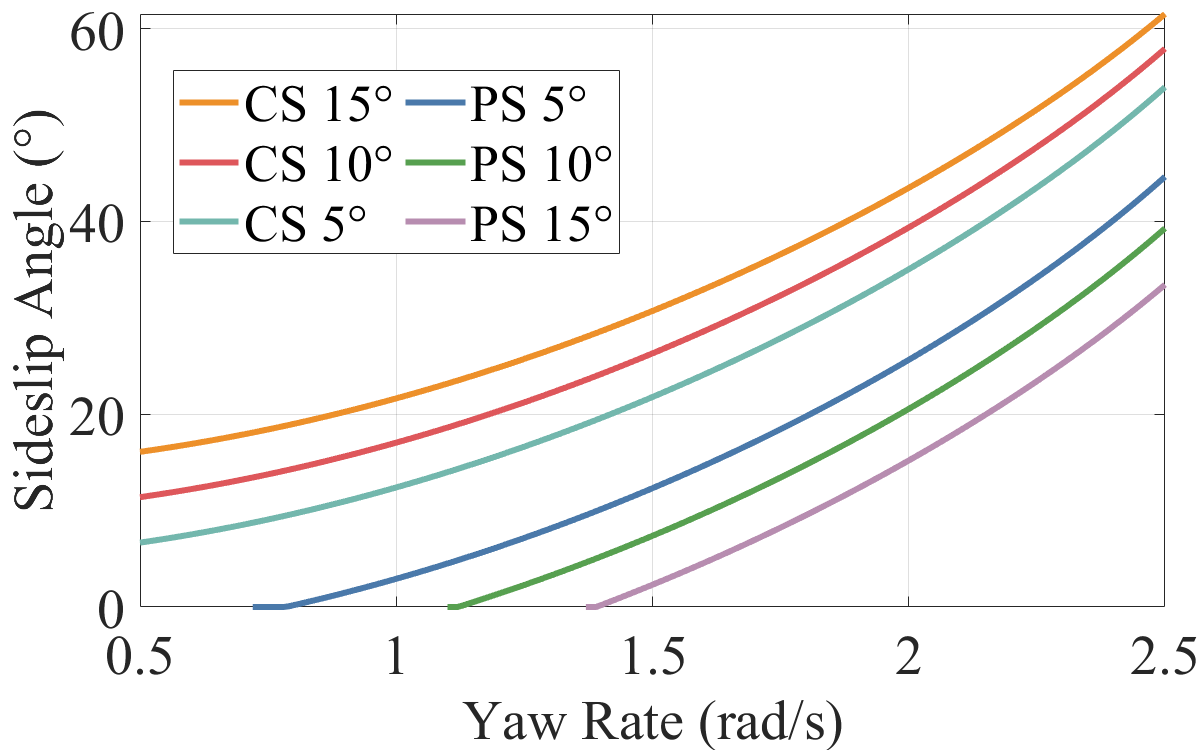}
    \caption{Rear wheel sideslip angles $\delta_r^{ss}$ of the steady-state drifting equilibrium.}
    \label{subfig:eq of sideslip ang}
    \end{subfigure} 
    ~
    \begin{subfigure}[t]{0.3 \textwidth}
    \includegraphics[width = 1.0 \textwidth]{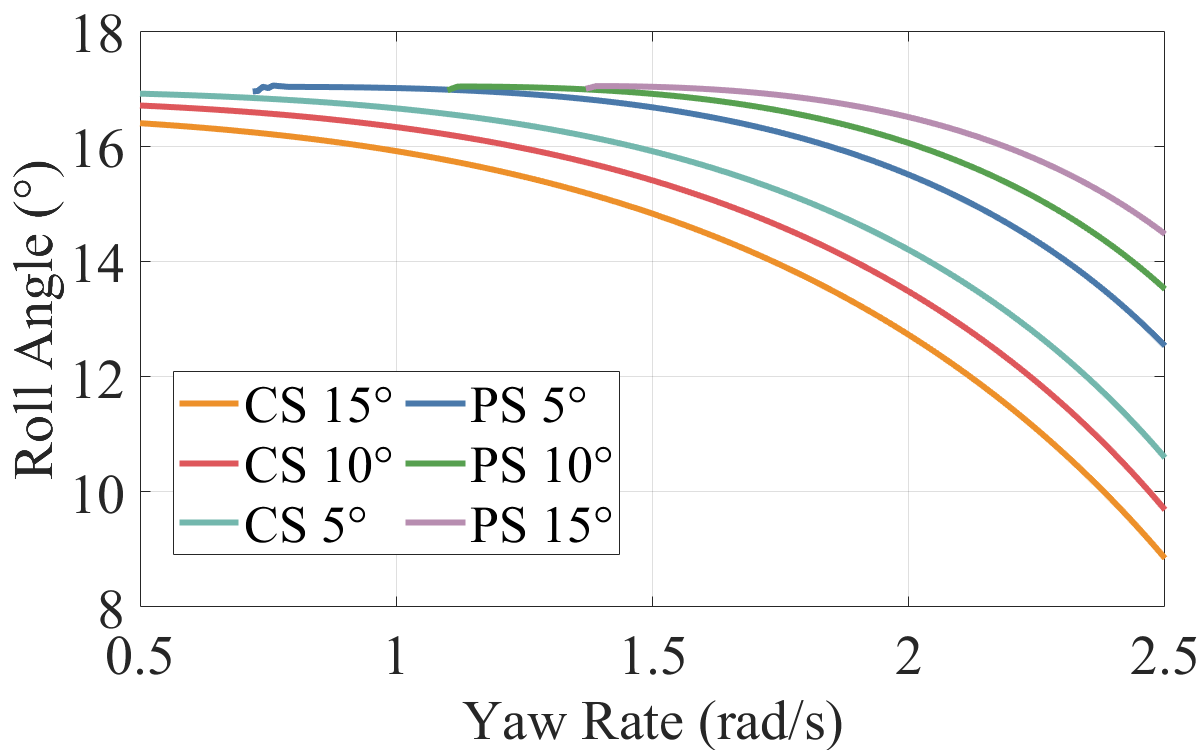}
    \caption{Roll angles $\varphi^{ss}$ of the steady-state drifting equilibrium.}
    \label{subfig:eq of roll ang}
    \end{subfigure}
     ~
    \begin{subfigure}[t]{0.3 \textwidth}
    \includegraphics[width = 1.0 \textwidth]{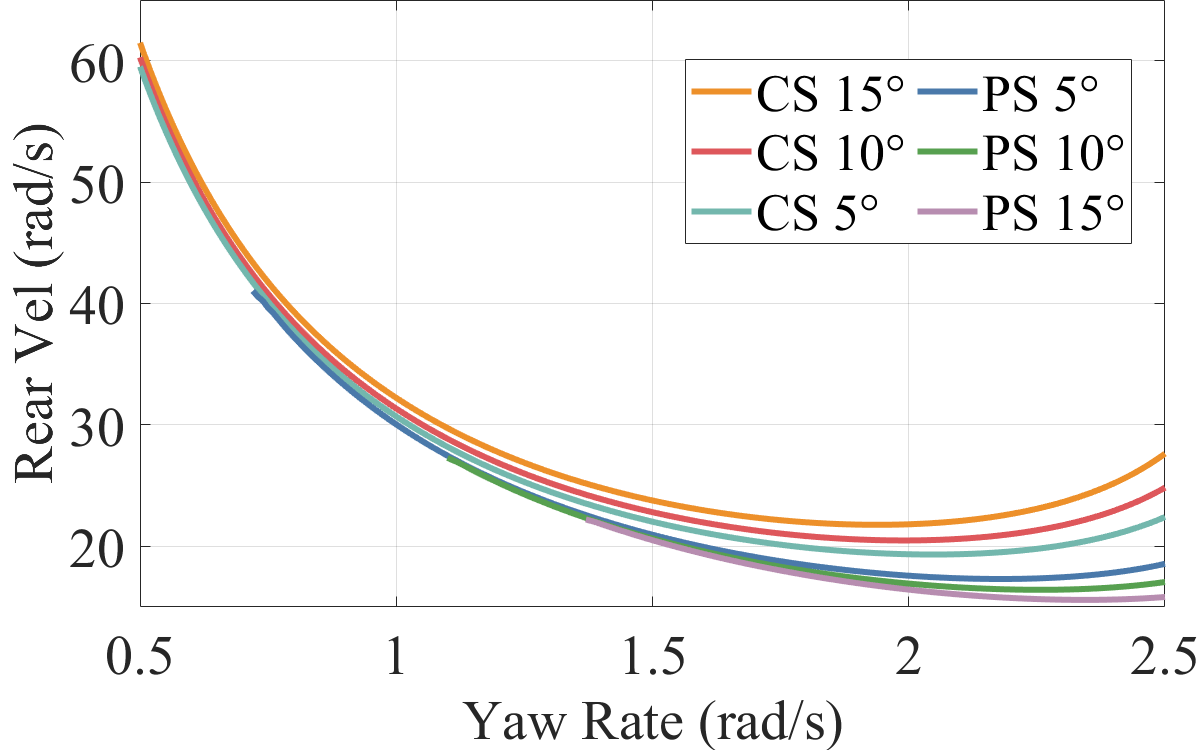}
    \caption{Rear angular velocities $\omega_r^{ss}$ of the steady-state drifting equilibrium.}
    \label{subfig:eq of rear vel}
    \end{subfigure}
    \caption{Steady-state drifting equilibrium under different yaw rates and steering angles. 'CS' stands for 'Counter-Steering', and 'PS' stands for 'Positive-Steering'.}
    \label{fig: eq curves}
\end{figure*}

As mentioned above, specifying two variables in $\bm{\xi}$ corresponds to a drifting equilibrium point. Here, we start from the projected steering angle $\delta_f$ (closely related to steering angle $\delta$) and the yaw rate $\dot{\psi}$ to compute the other unknown equilibrium point variables. With the longitudinal weight transfer neglected, the rear wheel friction $f_r$ is determined by
\begin{equation}
    f_r = \mu N_r \approx \mu mg\frac{b-a}{b}.
    \label{eq:fr}
\end{equation}

By considering (\ref{eq:Fg}) and (\ref{eq:fr}), the moment balance equation is expressed as 
\begin{align}
    f_r b\sin(\delta_r) &= F_G(b-a)\sin(\delta_G)  \nonumber\\
    \Rightarrow \qquad \mu g\sin(\delta_r) &= \dot{\psi}^2R_G\sin(\delta_G).
\end{align}

By applying the Law of Sines, the rear wheel contact point's radius of rotation $R_r$ and the rear wheel sideslip angle $\delta_r$ are determined by
\begin{align}
    R_r &= R_G\frac{\sin(\delta_G)}{\sin(\delta_r)}=\frac{\mu g}{\dot{\psi}^2},  \label{eq: Rr} \\
    \delta_r = \pi - (\pi/2+&\delta_f) - \arcsin(b\sin(\pi/2+\delta_f)/R_r). \label{eq: delta_r}
\end{align}

Then, the velocity of the rear wheel contact point and the angular velocity of the rear wheel are derived as
\begin{equation}
    v_r = -\dot{\psi}R_r,\qquad \omega_r =- \frac{v_r}{r\sin(\delta_r)},
    \label{eq: vr and wr}
\end{equation}
where the positive direction is determined by the right-hand rule. According to constraint (\ref{eq: constraint fx}), the angular velocity of the front wheel is
\begin{equation}
    \omega_f = -\frac{v_r\sin(\delta_r)}{r\cos(\delta_f)}.
    \label{eq: wf}
\end{equation}

The small angle approximations $\sin(\varphi) \approx \varphi$, 
 $\cos(\varphi) \approx 1$ are applied to the equation of roll motion (\ref{eq: eom varphi}). Then, the roll angle $\varphi$ is determined by
\begin{equation}
    \varphi = -\frac{I_r\omega_r\dot{\psi}-mv_r\sin(\delta_r)h\dot{\psi}}{mh^2\dot{\psi}^2 + mgh}.
    \label{eq: phi}
\end{equation}

Finally, the steering angle $\delta$ can be calculated by the geometry equation (\ref{eq: steering projection}), which is reformulated as
\begin{equation}
    \delta = \arctan\left(\frac{\tan(\delta_f)\cos(\varphi)}{\cos(\lambda)} \right).
    \label{eq: delta}
\end{equation}

\begin{algorithm}[t]
\caption{ADESA}
\label{alg: ADEA}
\begin{algorithmic}[1] 
\REQUIRE Steering angle $\delta^{ss}$, yaw rate $\dot{\psi}^{ss}$, tolerance $\epsilon$
\ENSURE Steady-state drifting equilibrium $\bm{\xi}^{ss}$
\STATE \textbf{initialize:} $\delta_f = \delta^{ss}$, $\dot{\psi}=\dot{\psi}^{ss}$, $e=\infty$, $\alpha=1$
\FOR {$e < \epsilon$}
    \STATE Solve $R_r$ and $\delta_r$ according to (\ref{eq: Rr}) and (\ref{eq: delta_r})
    \STATE Solve $\omega_r$, $\omega_f$, $\varphi$, and $\delta$ according to (\ref{eq: vr and wr})--(\ref{eq: delta})
    \STATE $e \leftarrow \delta-\delta^{ss}$
    \STATE $\delta_f \leftarrow \delta_f - \alpha \cdot e$
\ENDFOR

\RETURN $\bm{\xi}^{ss}=\begin{bmatrix}
    \delta & \varphi & \dot{\psi} & \omega_f & \omega_r
\end{bmatrix}^T$
\end{algorithmic}
\end{algorithm}

At this point, the equilibrium states $\bm{\xi}$ calculation is complete. However, the initial condition includes the projection steering angle $\delta_f$, an intermediate variable that is difficult to measure directly, which can be addressed by iterative algorithms. Then the analytical drifting equilibrium solving algorithm (ADESA) that computes the drifting equilibrium point starting from the steering angle $\delta$ and the yaw rate $\dot{\psi}$ is provided in Algorithm \ref{alg: ADEA}. The computational efficiency of ADESA compared to the numerical method is validated in Section V.

\subsection{Equilibrium Analysis}

In order to reveal the underlying drifting mechanisms of the STTW robots, the steady-state drifting equilibrium curves are calculated using the numerical method, as illustrated in Fig. \ref{fig: eq curves}. Each drifting equilibrium point is determined by the steering angle $\delta$ and the yaw rate $\dot{\psi}$. Counter-steering refers to steering in the opposite direction to the yaw rate, while positive-steering in the same direction. The physical parameters of the STTW robot are listed in Table \ref{table: params}. 

\begin{table}[h]
\caption{Physical parameter values of the STTW robot}
\begin{center}
\begin{tabular}{|c|c||c|c|}
\hline
Parameter & Value & Parameter & Value \\
\hline
$m$ (kg) & 5.435 & $a$ (m) & 0.164\\
\hline
$Ic_{xx}$ (kg$\cdot$m$^2$) & 3.31$\times 10^{-2}$ & $b$ (m) & 0.402 \\
\hline
$Ic_{zz}$ (kg$\cdot$m$^2$) & 9.40$\times 10^{-2}$ & $c$ (m) & 0.023 \\
\hline
$I_f$ (kg$\cdot$m$^2$) & 2.03$\times 10^{-2}$ & $h$ (m) & 0.195 \\
\hline
$I_r$ (kg$\cdot$m$^2$) & 2.17$\times 10^{-2}$ & $r$ (m) & 0.100 \\
\hline
$\mu$  & 0.3 & $\lambda$ ($^\circ$) & 25 \\
\hline
\end{tabular}
\end{center}
\label{table: params}
\end{table}

The most important feature shown in Fig. \ref{fig: eq curves}(a) is that for small yaw rates, the sideslip angle at the positive-steering drift equilibrium tends to zero, or no drift equilibrium point exists, which reveals the reason that professional motorcyclists do not adopt positive-steering drifting techniques. Another way to explain this phenomenon is that the geometry relationship shown in Fig. \ref{fig: simplified drifting model} indicates that the counter-steering increases the rear wheels' instantaneous turning radius, facilitating the transition into drifting.

By analyzing the trends in the drifting equilibrium point's curves, we can find several characteristics of STTW robots' drifting maneuver. 

(i) In Fig. \ref{fig: simplified drifting model}, with an increase in the counter-steering angle, the COM instantaneous turning radius $R_G$ also increases according to geometric relationships. Thus, (\ref{eq:Fg}) indicates that a higher counter-steering angle or yaw rate will lead to an increase of the inertial force $F_G$, which decreases the roll angle according to the moment balance function, the same as shown in Fig. \ref{fig: eq curves}(b). 

(ii) As the rear-wheel friction $f_r$ belongs to sliding friction with relatively stable value, the only way to maintain balance is to increase the sideslip angle $\delta_r$, as depicted in Fig. \ref{fig: eq curves}(a). 

(iii) According to (\ref{eq: vr and wr}), a larger sideslip angle corresponds to a lower rear wheel angular velocity $\omega_r$, which is consistent with the left side of Fig. \ref{fig: eq curves}(c). 

In summary, intrinsic dynamic relationships can elucidate most aspects of drifting behavior and offer a clearer insight into the drifting mechanism of STTW robots.

\section{Drifting Controller}
This section presents an MPC algorithm to perform two tasks. The first task is the steady-state drifting control. To formulate the MPC problem, we use the drifting model as the nominal model and the equilibrium points to construct the cost function. The second task involves transition between equilibrium points, where the ADESA method serves as a warm-start tool for the controller.
\subsection{Steady-state Drifting Control}
This task aims to stabilize the STTW robot at the drifting equilibrium point. Based on the drifting model and equilibrium analysis mentioned above, the control problem is formulated as the following optimization problem:
\begin{align} 
    \mathop{\min}_{\bm{x}_k,\bm{u}_k}& \quad 
    \sum\limits_{k=0}\limits^{N}  \| \bm{x}_k-\bm{x}^{ss} \|^2_{\bm{Q}} 
    + \| \bm{u}_k-\bm{u}^{ss} \|^2_{\bm{R}}
    \label{cost function}      \\
    s.t.& \quad \bm{x}_{k+1}=\bm{f}(\bm{x}_k,\bm{u}_k)  \label{discretized model} \\
    & \quad \bm{x}_0 = \bm{x}(t)   \\
    & \quad \bm{x}_{min} \leq \bm{x}_k \leq \bm{x}_{max} \\
    & \quad \bm{u}_{min} \leq \bm{u}_k \leq \bm{u}_{max},
\end{align}
where $\|\bm{x}\|^2_{\bm{Q}}$ refers to $\bm{x}^T\bm{Q}x$, $\bm{Q}$,$\bm{R}$ are the cost matrices, $\bm{x}_{min}$,$\bm{u}_{min}$,$\bm{x}_{max}$,$\bm{u}_{max}$ are the bounds of the state and input variables and $N$ is the planning horizon. 

The cost function (\ref{cost function}) encourages the robot to keep steady-state drifting with constant values of the state variables. The discrete model (\ref{discretized model}) is derived from the drifting dynamics model (\ref{eq: drifting model}). The optimization problem above is solved by the OCS2 \cite{OCS2} in an MPC scheme.

\subsection{Drifting Equilibrium Transition Control}
The former controller achieves the steady-state drifting, resulting in a circular path. However, the trajectory is fixed and the robot can only make small adjustments. To increase drifting flexibility, we further develop a drifting equilibrium transition controller that enables the robot to adjust its trajectory over a wide range during drifting.

\begin{figure}[t]
  \centering{\includegraphics[width=1\linewidth]{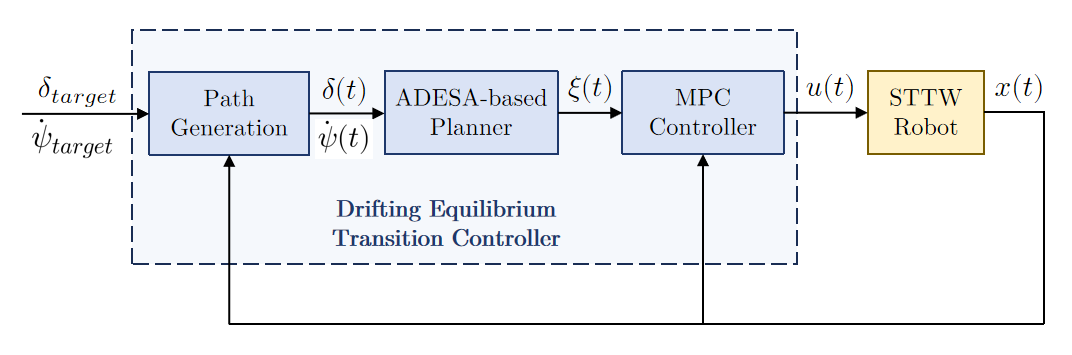}}
  \caption{Block diagram of the drifting equilibrium transition controller.}
  \label{fig: state trans controller}
\end{figure}

Fig. \ref{fig: state trans controller} shows the block diagram of the drifting equilibrium transition controller. $\delta_{target}$ and $\dot{\psi}_{target}$ are the target steering angle and the yaw rate. The trajectory curves $\delta(t)$ and $\dot{\psi}(t)$ are generated through interpolation methods. As stated before, an equilibrium point can be determined by two variables. Thus, by specifying the steering angle $\delta(t)$ and the yaw rate $\dot{\psi}(t)$, the ADESA method can be used to calculate the other state trajectories, forming a trajectory composed of equilibrium points. However, this trajectory is not dynamically feasible, but it can serve as a good initialization for the MPC algorithm to run until convergence. The cost funtion of the controller is modified as:
\begin{align}
    \mathop{\min}_{\bm{x}_k,\bm{u}_k} &\quad 
    \| \bm{x}_N-\bm{x}^{ss}_{target} \|^2_{\bm{Q}_f} \nonumber\\    
    & \quad\qquad +\sum\limits_{k=0}\limits^{N-1}  \| \bm{x}_k-\bm{x}_k^{ss} \|^2_{\bm{Q}} 
    + \| \bm{u}_k-\bm{u}_k^{ss} \|^2_{\bm{R}}.
    \label{cost function1} 
\end{align} 

This cost function is designed to guide the robot drifting along the trajectory planned by the ADESA method.
When MPC runs at a high rate, the optimization results generally do not vary significantly. Therefore, the previous trajectory is a good initialization for the next step. However, in the drifting equilibrium transition task, the former optimization result does not provide enough guidance for trajectory modification. 
Thus, ADESA plays an important role in this task by quickly generating a nearly feasible trajectory to warm-start the MPC loop.

\section{Simulation Results}

In this section, we conduct simulation experiments to draw the following conclusions.

(i) By comparing the results of the NDESA and ADESA with the steady-state drifting data obtained from simulations, the drifting equilibrium solving algorithms are accurate within an acceptable error range.

(ii) Through equilibrium transition simulations, the MPC controller guided by the ADSEA method enables wide-range equilibrium point switching.

(iii) Tested by drifting simulations on variable-friction surfaces, the proposed controller is robust under model parameter perturbations, suggesting its potential for real-world applications.

\begin{figure*}[t]
  \centering
    \begin{subfigure}[t]{0.23 \textwidth}
    \includegraphics[width = 1.0 \textwidth]{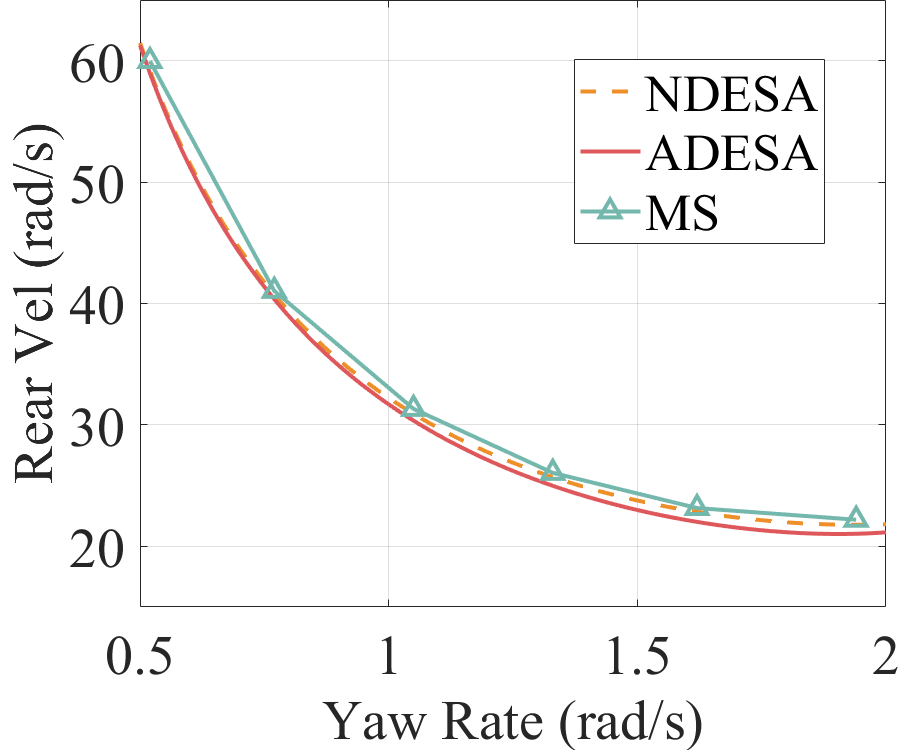}
    \caption{Rear angular velocities.}
    \label{subfig:compare of rear vel}
    \end{subfigure} 
    ~
    \begin{subfigure}[t]{0.23 \textwidth}
    \includegraphics[width = 1.0 \textwidth]{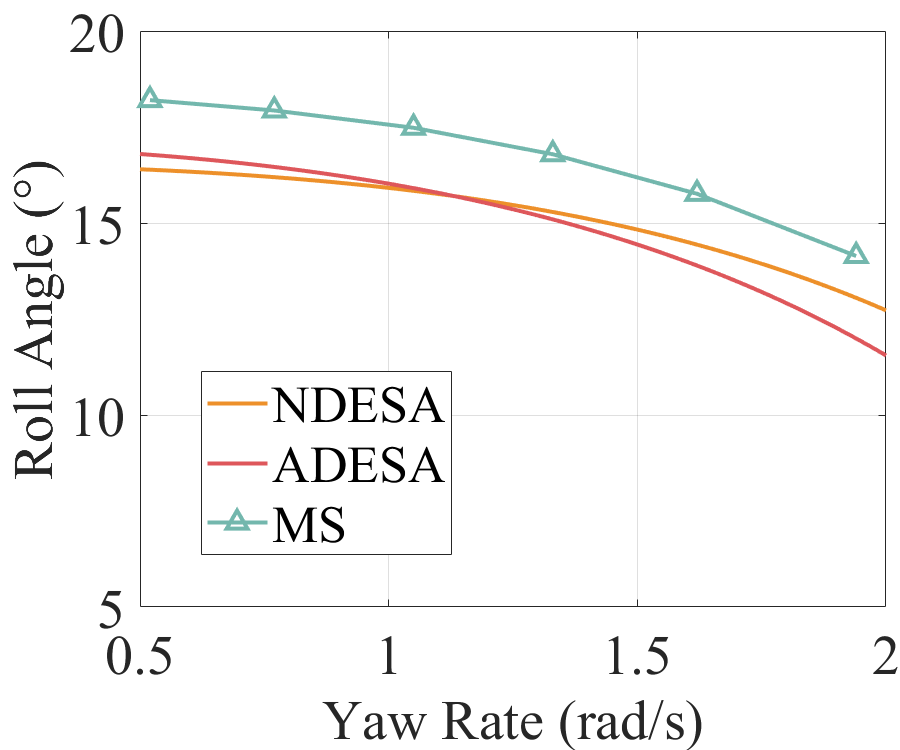}
    \caption{Roll angles.}
    \label{subfig:compare of roll ang}
    \end{subfigure}
    ~
    \begin{subfigure}[t]{0.23 \textwidth}
    \includegraphics[width = 1.0 \textwidth]{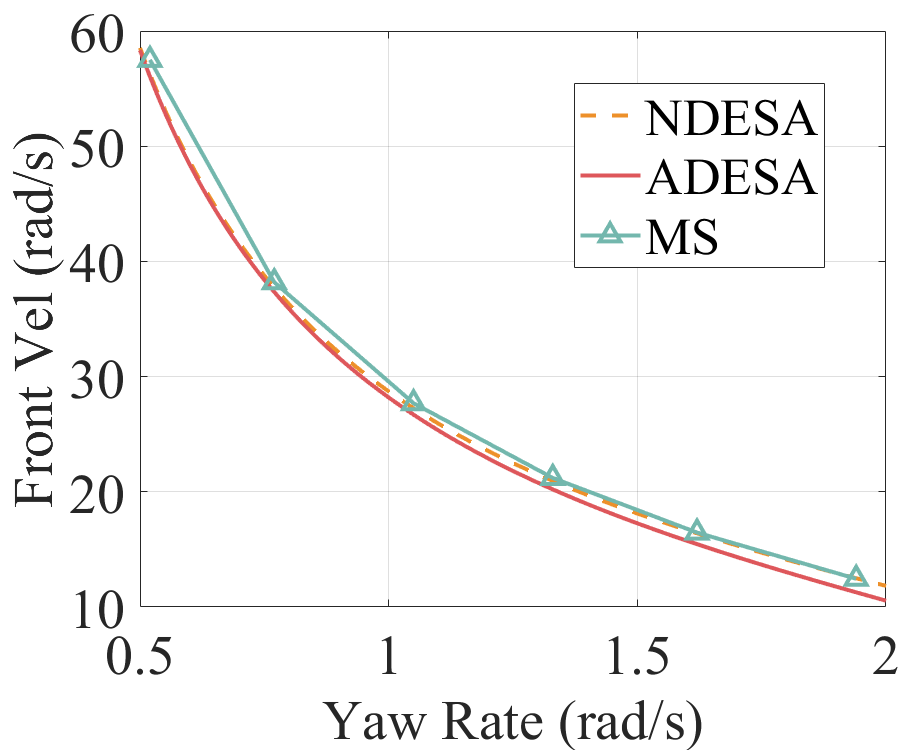}
    \caption{Front angular velocities.}
    \label{subfig:compare of front vel}
    \end{subfigure}
    ~
    \begin{subfigure}[t]{0.23 \textwidth}
    \includegraphics[width = 1.0 \textwidth]{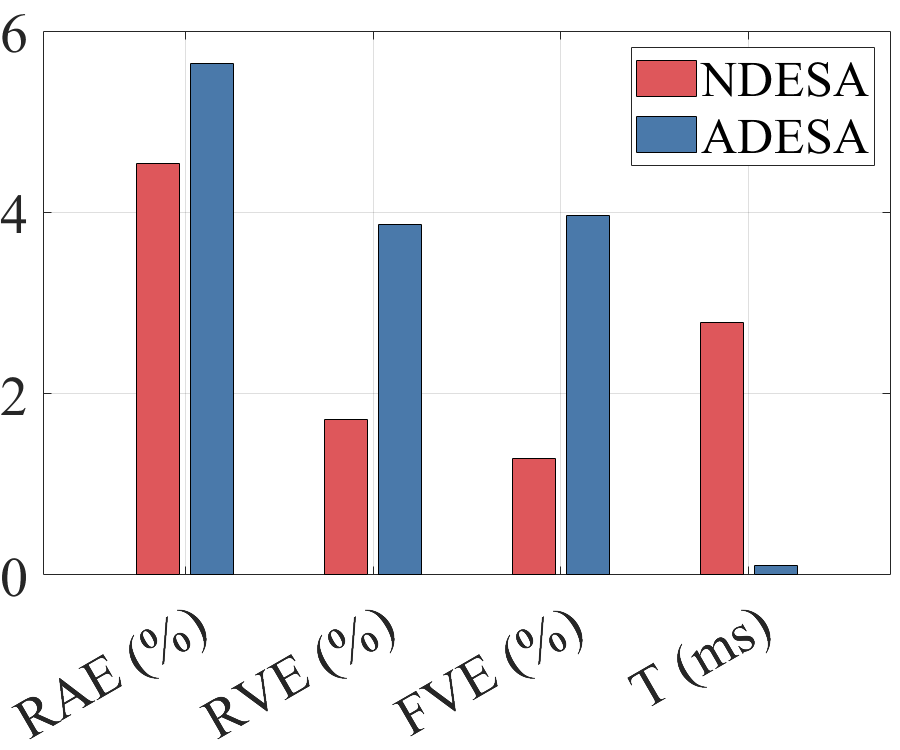}
    \caption{Percentage relative error and computation time.}
    \label{subfig:bar}
    \end{subfigure}
    \caption{Comparisons of equilibrium accuracy and computational time between the numerical drifting equilibrium solving algorithm (NDESA) and analytical drifting equilibrium solving algorithm (ADESA). `MS' represents MuJoCo simulations, considered to be true value of the drifting equilibrium points. In (d), `RAE', `RVE', and `FVE' represent `roll angle error', `rear wheel velocity error', and `front wheel velocity error', respectively. `T' is the computation time.}
    \label{fig: compare}
\end{figure*}

\begin{figure}[t]
  \centering{\includegraphics[width=0.98\linewidth]{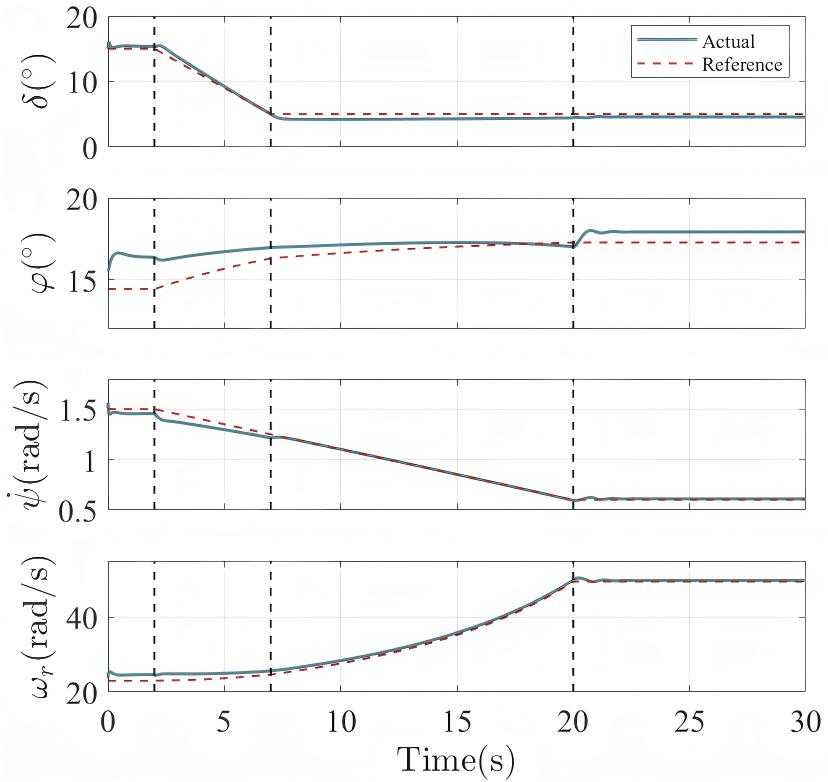}}
  \caption{State trajectories with the drifting equilibrium transition controller. The red dashed lines are reference trajectories, calculated through the ADESA method. The blue solid lines represent the actual tracking result.}
  \label{fig: state trans}
\end{figure}

\begin{figure}[t]
  \centering{\includegraphics[width=1\linewidth]{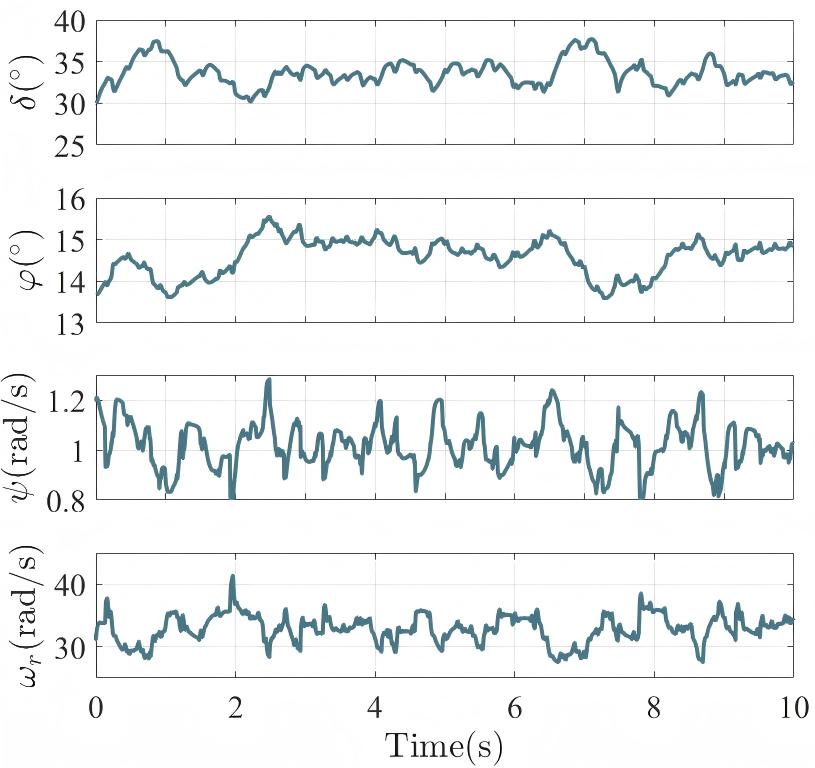}}
  \caption{State trajectories of the robot drifting on variable-friction surfaces. }
  \label{fig: mujoco noise}
\end{figure}

\subsection{Comparison of Drifting Equilibrium}
An STTW robot is built in the MuJoCo simulation environment with the parameters shown in Table \ref{table: params}. By setting the initial state of the STTW robot near the drifting equilibrium point in the simulation, the proposed MPC controller in Section IV achieves steady-state drifting control, as shown in Fig. \ref{fig: cyclist and mujoco simulation}, and the corresponding video is provided in the supplementary material.

The success of steady-state drifting control in simulation can help validate the accuracy of the equilibrium solving algorithms. Figs. \ref{fig: compare}(a)--(c) show the variation of the drift equilibrium point with yaw rate when the robot's steering is held at a counter-steering angle of 15$^\circ$, and Fig. \ref{fig: compare}(d) summarizes the comparison of equilibrium points with steering angle controlled at 5$^\circ$, 10$^\circ$ and 15$^\circ$. Based on the comparison results, the numerical solving algorithm demonstrates an average error of roughly 4\%, with each solution taking around 2.8 ms, while the ADESA algorithm has a slightly higher average error of 6\%, yet achieves a much faster solving time of just 0.3 $\mu\mathrm{s}$. All computations of the drifting equilibrium were performed using MATLAB R2021b on a personal computer equipped with an 11th Gen Intel(R) Core(TM) i5-1155G7 CPU running at 2.50 GHz.



\subsection{Drifting Equilibrium Transition Control}
In the drifting equilibrium transition simulation, the steering angle gradually decreases from 15$^\circ$ at 2s and remains at 5$^\circ$ starting from 7s, while the yaw rate decreases from 1.5 rad/s at 2s and remains at 0.6 rad/s from 20s. Leveraging the fast-solving characteristics of the ADESA method, the MPC controller is warm-started to achieve the switch between equilibrium points, as shown in Fig. \ref{fig: state trans}. 
The results show a large roll angle tracking error due to approximations in the ADESA method, as demonstrated earlier. Larger errors at 2s, 7s, and 20s arise because the ADESA-planned trajectory comprises equilibrium points that are not dynamically feasible. Nonetheless, their proximity to the targets allows the controller to achieve equilibrium transition task.

\subsection{Drifting on Variable-friction Surfaces}
Accurate friction is essential for robot drifting, but it remains difficult to measure in real-world scenarios. Therefore, we explore the controller's robustness against friction perturbations. In this experiment, the ground is composed of 0.5m × 0.5m blocks, each with a friction coefficient randomly selected between 0.25 and 0.35. This variable-friction ground is shown in the third part of the supplementary video. In contrast, the MPC controller's nominal model assumes a constant friction coefficient of 0.3. Although there is a friction mismatch between the nominal and simulation model, the controller can still achieve steady-state drifting as shown in Fig. \ref{fig: mujoco noise}. In the future, we will further integrate a friction identification algorithm to enhance the robustness of drifting control.

\section{Conclusions}

In this paper, we extend the concept of drifting to intrinsically unstable STTW robots, in which the additional DOF in the roll direction presents a significant challenge for this task. By establishing a drifting dynamics model with large sideslip angle, the drifting equilibrium theory is developed for STTW robots. The underlying drifting mechanism, especially the counter-steering drifting technique is explained through the equilibrium solution. In addition to numerical method for solving drifting equilibrium, an analytical algorithm ADESA is proposed to speed up the process by orders of magnitude with a slight loss in accuracy. The successful steady-state drifting and equilibrium transition control in simulation further supports the reliability of the dynamic model.

Though elucidating the drift mechanism is the basic work for achieving drifting, there remains a considerable gaps to real-world application, and we aim to achieve real-world drifting of STTW robots in the future.








\bibliographystyle{IEEEtran}
\bibliography{ref}

\end{document}